\documentclass[letterpaper]{article}
\usepackage[preprint]{aaai2027}
\usepackage[hyphens]{url}
\usepackage{graphicx}
\usepackage{natbib}
\usepackage{caption}
\usepackage{amsmath}
\usepackage{amssymb}
\usepackage{booktabs}

\newcommand{\gclean}{\mathbf{g}_{\mathrm{clean}}}
\newcommand{\gadv}{\mathbf{g}_{\mathrm{adv}}}
\newcommand{\gawp}{\mathbf{g}_{\mathrm{awp}}}
\newcommand{\one}{\mathbf{1}}
\newcommand{\argmax}{\operatorname*{arg\,max}}
\newcommand{\Dmat}{\mathbf{D}}
\newcommand{\betavec}{\boldsymbol{\beta}}
\newcommand{\gbeta}{\mathbf{g}_{\betavec}}
\newcommand{\avec}{\mathbf{a}}
\newcommand{\TableValue}[2]{#1\(\pm\)#2}
\newcommand{\TableBest}[2]{\textbf{#1}\(\pm\)\textbf{#2}}
\newcommand{\TableSecond}[2]{\underline{#1\(\pm\)#2}}

\newcommand{\ExGradOverviewFigure}{%
\begin{figure*}[t]
\centering
\includegraphics[width=\textwidth]{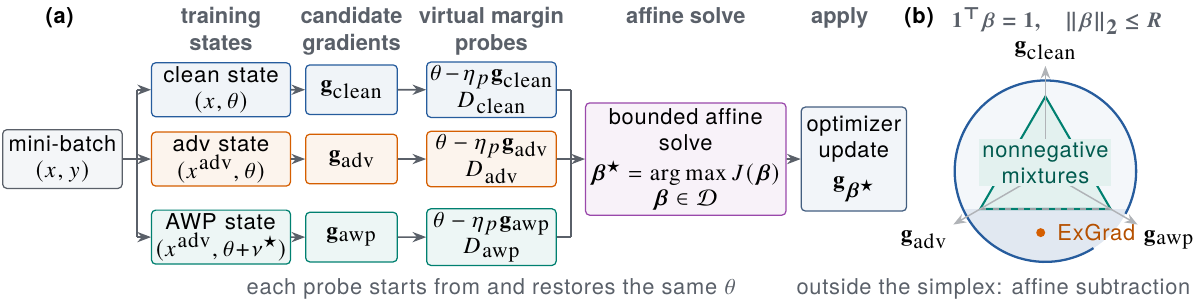}
\caption{Overview of ExGrad.
(a) A mini-batch update in which three state-dependent gradients are probed
from the same nominal parameters; their margin responses determine one affine
recomposition, and only the recomposed gradient advances the optimizer.
(b) The bounded unit-sum domain intersects the nonnegative simplex and extends
beyond it to permit response-supported subtraction.}
\label{fig:overview}
\end{figure*}
}

\newcommand{\ExGradDynamicsFigure}{%
\begin{figure}[t]
\centering
\includegraphics[width=\columnwidth]{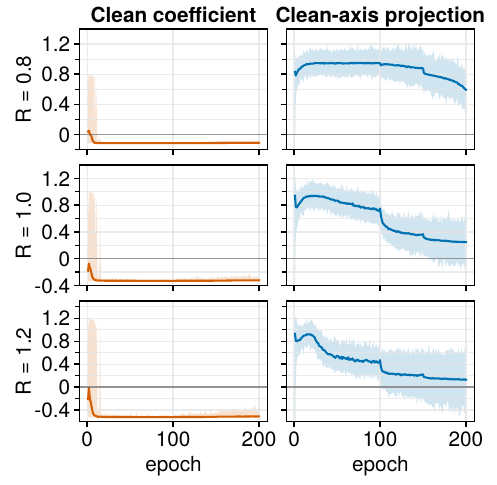}
\caption{Clean-coordinate and clean-axis projection dynamics on CIFAR-10
with WRN-28-4. Rows correspond to $R=0.8$, $1.0$, and $1.2$; the left and
right columns show
$\beta_{\mathrm{clean}}$ and $p_{\mathrm{clean}}$, respectively. Solid curves
are within-epoch mini-batch means; shaded regions span the mini-batch minima
and maxima.}
\label{fig:dynamics}
\end{figure}
}

\newcommand{\ExGradTrainingCurvesFigure}{%
\begin{figure}[t]
\centering
\includegraphics[width=\columnwidth]{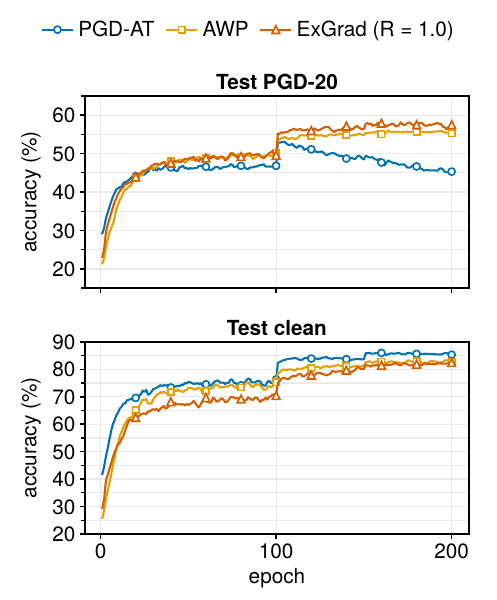}
\caption{Test-accuracy trajectories on CIFAR-10 with WRN-28-4. The upper and
lower panels report PGD-20 and clean accuracy, respectively; ExGrad uses
$R=1.0$. All trajectories use a three-epoch moving average.}
\label{fig:training-curves}
\end{figure}
}

\newcommand{\ExGradMainResultsTable}{%
\begin{table*}[t]
\centering
\normalsize
\setlength{\tabcolsep}{2.0pt}
\renewcommand{\arraystretch}{1.05}
\begin{tabular*}{\textwidth}{@{\extracolsep{\fill}}lllccccc@{}}
\toprule
Dataset & Model & Method & Clean & PGD & APGD-CE & APGD-T & AA \\
\midrule
CIFAR-10 & WRN-28-4 & PGD-AT & \TableValue{83.00}{0.47} & \TableValue{53.41}{0.28} & \TableValue{52.69}{0.27} & \TableValue{49.15}{0.21} & \TableValue{49.12}{0.21} \\
 &  & TRADES ($\beta=3$) & \TableBest{84.94}{1.84} & \TableValue{50.23}{0.39} & \TableValue{49.40}{0.41} & \TableValue{47.50}{0.33} & \TableValue{47.51}{0.33} \\
 &  & TRADES ($\beta=6$) & \TableSecond{84.21}{0.67} & \TableValue{52.33}{0.38} & \TableValue{51.65}{0.39} & \TableValue{49.54}{0.08} & \TableValue{49.53}{0.05} \\
 &  & TRADES ($\beta=9$) & \TableValue{82.40}{0.30} & \TableValue{53.12}{0.23} & \TableValue{52.38}{0.19} & \TableValue{49.89}{0.15} & \TableValue{49.88}{0.16} \\
 &  & AWP & \TableValue{82.62}{0.34} & \TableValue{56.48}{0.22} & \TableValue{55.80}{0.27} & \TableSecond{51.09}{0.09} & \TableSecond{51.07}{0.11} \\
 &  & ExGrad ($R=0.8$) & \TableValue{83.75}{0.21} & \TableValue{56.51}{0.17} & \TableValue{55.57}{0.19} & \TableBest{51.25}{0.31} & \TableBest{51.25}{0.30} \\
 &  & ExGrad ($R=1.0$) & \TableValue{81.19}{0.23} & \TableSecond{58.95}{0.09} & \TableBest{57.64}{0.12} & \TableValue{50.72}{0.04} & \TableValue{50.71}{0.03} \\
 &  & ExGrad ($R=1.2$) & \TableValue{79.77}{0.08} & \TableBest{59.23}{0.27} & \TableSecond{57.48}{0.17} & \TableValue{49.98}{0.51} & \TableValue{49.96}{0.48} \\
\midrule
CIFAR-10 & WRN-34-10 & PGD-AT & \TableSecond{85.74}{0.71} & \TableValue{55.21}{0.38} & \TableValue{54.49}{0.49} & \TableValue{51.86}{0.31} & \TableValue{51.84}{0.33} \\
 &  & AWP & \TableBest{86.13}{0.22} & \TableValue{58.52}{0.18} & \TableValue{57.63}{0.21} & \TableSecond{53.49}{0.38} & \TableSecond{53.51}{0.39} \\
 &  & ExGrad ($R=0.8$) & \TableValue{85.30}{0.12} & \TableValue{58.97}{0.22} & \TableValue{58.22}{0.26} & \TableBest{53.97}{0.23} & \TableBest{53.98}{0.23} \\
 &  & ExGrad ($R=1.0$) & \TableValue{84.58}{0.87} & \TableBest{61.77}{0.37} & \TableBest{60.28}{0.30} & \TableValue{52.62}{0.19} & \TableValue{52.65}{0.19} \\
 &  & ExGrad ($R=1.2$) & \TableValue{83.27}{0.50} & \TableSecond{61.21}{0.11} & \TableSecond{59.12}{0.29} & \TableValue{51.02}{0.45} & \TableValue{51.00}{0.47} \\
\midrule
CIFAR-100 & WRN-28-4 & PGD-AT & \TableValue{53.69}{1.37} & \TableValue{28.82}{0.26} & \TableValue{28.23}{0.32} & \TableValue{24.43}{0.28} & \TableValue{24.42}{0.29} \\
 &  & AWP & \TableValue{53.96}{0.21} & \TableValue{31.28}{0.11} & \TableValue{30.83}{0.02} & \TableValue{25.82}{0.24} & \TableValue{25.78}{0.23} \\
 &  & ExGrad ($R=0.8$) & \TableBest{57.80}{0.63} & \TableSecond{32.71}{0.16} & \TableSecond{31.95}{0.06} & \TableSecond{27.54}{0.20} & \TableSecond{27.53}{0.19} \\
 &  & ExGrad ($R=1.0$) & \TableSecond{54.32}{0.24} & \TableBest{32.99}{0.25} & \TableBest{32.34}{0.28} & \TableBest{27.64}{0.27} & \TableBest{27.63}{0.27} \\
 &  & ExGrad ($R=1.2$) & \TableValue{50.70}{0.54} & \TableValue{32.49}{0.14} & \TableValue{31.79}{0.05} & \TableValue{26.79}{0.17} & \TableValue{26.77}{0.19} \\
\midrule
ImageNet-100 & WRN-28-4 & PGD-AT & \TableBest{75.86}{0.31} & \TableValue{49.31}{0.34} & \TableValue{48.31}{0.29} & \TableValue{44.82}{0.26} & \TableValue{44.81}{0.25} \\
 &  & AWP & \TableValue{73.02}{1.90} & \TableValue{49.09}{0.79} & \TableValue{48.14}{0.82} & \TableValue{43.24}{1.18} & \TableValue{43.21}{1.21} \\
 &  & ExGrad ($R=0.8$) & \TableSecond{73.53}{0.42} & \TableBest{50.81}{0.30} & \TableBest{49.93}{0.36} & \TableSecond{45.38}{0.55} & \TableSecond{45.36}{0.57} \\
 &  & ExGrad ($R=1.0$) & \TableValue{71.24}{0.12} & \TableSecond{50.61}{0.57} & \TableSecond{49.89}{0.57} & \TableBest{45.58}{0.69} & \TableBest{45.57}{0.69} \\
 &  & ExGrad ($R=1.2$) & \TableValue{69.07}{0.32} & \TableValue{50.04}{0.43} & \TableValue{49.39}{0.30} & \TableValue{44.57}{0.25} & \TableValue{44.57}{0.23} \\
\bottomrule
\end{tabular*}
\caption{Clean and robust test accuracy (\%). Best and second-best results
within each dataset--model block are boldfaced and underlined, respectively.}
\label{tab:main-results}
\end{table*}
}

\newcommand{\ExGradInterventionTable}{%
\begin{table}[t]
\centering
\normalsize
\setlength{\tabcolsep}{2.0pt}
\renewcommand{\arraystretch}{1.06}
\begin{tabular*}{\columnwidth}{@{\extracolsep{\fill}}ccccc@{}}
\toprule
$\beta_{\mathrm{clean}}$ & $\beta_{\mathrm{adv}}$ &
$\beta_{\mathrm{awp}}$ & Clean & PGD \\
\midrule
Adap.  & Adap. & $0$   & \TableValue{83.38}{0.39} & \TableValue{53.37}{0.38} \\
Adap.  & $0$   & Adap. & \TableValue{82.75}{0.37} & \TableValue{56.39}{0.16} \\
$0.1$  & Adap. & Adap. & \TableBest{84.91}{0.49} & \TableValue{55.53}{0.23} \\
$0$    & Adap. & Adap. & \TableValue{83.27}{0.33} & \TableValue{56.44}{0.10} \\
$-0.1$ & Adap. & Adap. & \TableValue{81.81}{0.14} & \TableValue{57.12}{0.14} \\
$-0.2$ & Adap. & Adap. & \TableValue{81.17}{0.15} & \TableValue{57.40}{0.04} \\
$-0.326$ & $0.601$ & $0.725$ & \TableValue{80.85}{0.20} & \TableValue{58.35}{0.03} \\
Adap.  & Adap. & Adap. & \TableValue{81.19}{0.23} &
\TableBest{58.95}{0.09} \\
\bottomrule
\end{tabular*}
\caption{Ablations on CIFAR-10 with WRN-28-4. All settings use $R=1$.
``Adap.'' denotes an adaptive coefficient. The fixed triplet averages the
coefficients at the best-PGD checkpoint of each seed and is rounded to three
decimals.}
\label{tab:fixed-clean}
\end{table}
}

\newcommand{\ExGradNormControlTable}{%
\begin{table}[t]
\centering
\normalsize
\setlength{\tabcolsep}{1.0pt}
\renewcommand{\arraystretch}{1.06}
\begin{tabular*}{\columnwidth}{@{\extracolsep{\fill}}ccccc@{}}
\toprule
$R$ & Match & Clean & PGD & Norm \\
\midrule
0.8 & No  & \TableValue{83.75}{0.21} & \TableValue{56.51}{0.17} & 0.943 \\
0.8 & Yes & \TableValue{83.81}{0.29} & \TableBest{56.58}{0.26} & 1.000 \\
\midrule
1.0 & No  & \TableValue{81.19}{0.23} & \TableBest{58.95}{0.09} & 1.083 \\
1.0 & Yes & \TableValue{81.13}{0.26} & \TableValue{58.65}{0.27} & 1.000 \\
\midrule
1.2 & No  & \TableValue{79.77}{0.08} & \TableBest{59.23}{0.27} & 1.217 \\
1.2 & Yes & \TableValue{79.92}{0.46} & \TableValue{58.86}{0.12} & 1.000 \\
\bottomrule
\end{tabular*}
\caption{AWP-norm-matched control on CIFAR-10 with WRN-28-4. Norm is the
applied update-to-AWP norm ratio and is unity by construction after matching.}
\label{tab:norm-control}
\end{table}
}

\newcommand{\ExGradTrainingCostTable}{%
\begin{table}[ht]
\centering
\normalsize
\setlength{\tabcolsep}{2.0pt}
\renewcommand{\arraystretch}{1.05}
\begin{tabular*}{\columnwidth}{@{\extracolsep{\fill}}lrrrr@{}}
\toprule
Method & Time/epoch & Rel.\ & Alloc.\ & Reserv.\ \\
 & (s) & time & (MiB) & (MiB) \\
\midrule
PGD-AT & 44.20 & $0.84\times$ & 1096.90 & 1248 \\
AWP    & 52.45 & $1.00\times$ & 1149.26 & 1258 \\
ExGrad & 76.09 & $1.45\times$ & 1173.50 & 1302 \\
\bottomrule
\end{tabular*}
\caption{Diagnostic-free training cost measured in separate processes on one
H100 80GB (CIFAR-10/WRN-28-4, batch 128, FP32). Time averages synchronized
epochs 2--5 after warm-up; memory reports PyTorch peaks.}
\label{tab:training-cost}
\end{table}
}

\title{ExGrad: Beyond Suppression---Explicit Clean-Gradient Subtraction in Adversarial Training}

\author{Zhiyuan Ye\textsuperscript{1}, Xiangyu Zhou\textsuperscript{2},
Ji Qi\textsuperscript{2,*}, Hao Zhang\textsuperscript{1,*}, and Yi Zhou\textsuperscript{1,*}}
\affiliations{
\textsuperscript{1}University of Science and Technology of China\\
\textsuperscript{2}China Mobile (Suzhou) Software Technology Co., Ltd.\\
zhiyuan.ye@mail.ustc.edu.cn, \{zhouxiangyu,qiji\}@cms.chinamobile.com,\\
\{hzhangzy,yi\_zhou\}@ustc.edu.cn\\
*Corresponding authors
}

\begin{document}
\maketitle

\begin{abstract}
Adversarial training usually derives updates from predefined surrogate
losses, but source-loss descent does not directly characterize an
update's finite-step effect on current adversarial margins. We introduce
ExGrad (\emph{extrapolative gradient recomposition}), which constructs
gradients at clean-input, adversarial-input, and adversarial-weight-perturbed
(AWP) states for each mini-batch. Lightweight virtual steps measure their local
margin responses, after which a globally solvable
three-variable problem selects their coefficients over a bounded affine
domain without nonnegativity constraints. Across different affine radii $R$
on CIFAR-10, negative clean coefficients rapidly become predominant, and
their magnitude increases with $R$. Cross-zero
interventions show that setting this coefficient to zero is insufficient:
mild subtraction further improves PGD-20, while stronger subtraction
trades clean accuracy for additional PGD-20. In CIFAR-10 WRN-28-4 controls,
none of the three tested candidate-removal variants matches full ExGrad,
showing that subtraction requires both adversarial directions in this setting. Projection
diagnostics show that a negative clean coordinate usually coexists with
a positive net clean-axis projection, although both decrease as the
radius grows. Norm-matched controls further indicate that update
magnitude alone does not explain the moderate-radius gains. Experiments
on CIFAR-10, CIFAR-100, and ImageNet-100 show radius-dependent
improvements over AWP. On CIFAR-100, ExGrad with $R=0.8$ improves all
five reported metrics, including clean and AutoAttack accuracy by $3.84$
and $1.75$ percentage points, respectively. On ImageNet-100, the same
radius improves all four robust metrics over both AWP and PGD-AT. These
results identify explicit clean-gradient subtraction as a
response-revealed update preference whose strength and clean--robust
operating point are controlled by the affine radius.
\end{abstract}

\section{Introduction}

Adversarial robustness depends on the stability of a classifier's local
decision geometry. For an input $x$ with label $y$, define the signed
classification margin
\begin{equation*}
m_\theta(x,y)
=
f_y(x;\theta)-\max_{k\ne y}f_k(x;\theta),
\end{equation*}
where $f_k(x;\theta)$ is the logit of class $k$. A robust model must preserve a
sufficient margin not only at the clean input but throughout the allowed input
neighborhood \cite{szegedy2014intriguing,goodfellow2015explaining,madry2018towards}.
Decision-boundary distance and local loss linearity provide complementary
geometric views of this neighborhood
\cite{moosavi2016deepfool,qin2019local}.
Adversarial training can therefore be viewed as repeatedly reshaping this
local geometry through parameter updates.

Most robust-training procedures obtain an update directly from a predefined
surrogate loss. A gradient $\mathbf{g}=\nabla_\theta L(\theta)$ is the first-order
sensitivity of its source loss: moving along $-\mathbf{g}$ locally reduces $L$.
This source-loss descent sensitivity, however, is not the same quantity as the
response of a target adversarial margin. For a small step,
\begin{equation*}
\begin{aligned}
m_{\theta'}(x,y)-m_\theta(x,y)
&\approx
-\eta\langle\nabla_\theta m_\theta(x,y),\mathbf{g}\rangle,\\[-1pt]
\theta'&=\theta-\eta\mathbf{g}.
\end{aligned}
\end{equation*}
A direction may reduce the loss from which it was derived yet provide limited
margin improvement, or even compress the margin of some adversarial examples.
The locally preferred update is consequently determined by its effect on the
target geometry, not by its source label alone.

This mismatch is not resolved by tuning a surrogate-loss coefficient. Such a
coefficient controls how source objectives generate an update, but it does not
report how that update changes the adversarial margin of the current
mini-batch. Gradient norms and pairwise angles are also insufficient for this
purpose: they describe the candidate directions relative to one another,
rather than their effects on the target geometry. This motivates treating a
gradient as a candidate parameter intervention whose consequence should be
measured before it is selected.

This distinction becomes actionable when several training states already
provide plausible directions. We consider gradients evaluated on clean inputs,
adversarial inputs, and adversarial inputs under adversarial weight
perturbation (AWP), denoted by $\gclean$, $\gadv$, and $\gawp$.
The first captures nominal fitting, the second input-neighborhood robust
fitting, and the third the correction induced by a perturbed parameter
neighborhood \cite{wu2020adversarial}. Their effects can complement or cancel
one another and can change with the mini-batch and training stage. Fixed loss
weights, gradient norms, or pairwise angles do not directly measure these
effects on the adversarial margin.

These candidates are not three unrelated tasks. They form a nested sequence
of training states: clean input at nominal weights, adversarial input at
nominal weights, and adversarial input at perturbed weights. Each transition
adds a robustness-relevant state change, while the adversarial margin supplies
a common criterion for comparing all three candidate updates.

We introduce ExGrad, short for \emph{extrapolative gradient recomposition}.
For every mini-batch, ExGrad applies a lightweight virtual update along each
candidate gradient from the same nominal parameters, measures the resulting
finite-step adversarial-margin response, restores the parameters, and solves
for a single recomposed update. The coefficients satisfy
\begin{equation*}
\left\{
\begin{aligned}
\gbeta
&=
\beta_{\mathrm{clean}}\gclean
+\beta_{\mathrm{adv}}\gadv
+\beta_{\mathrm{awp}}\gawp,\\
\one^\top\betavec&=1,\qquad \|\betavec\|_2\le R.
\end{aligned}
\right.
\end{equation*}
The unit-sum constraint defines an affine combination, while the radius $R$
bounds its departure from balanced recomposition. Unlike a nonnegative loss
mixture, this domain lets the measured response suppress, remove, or subtract
a candidate direction. The coefficients are therefore determined from the
measured responses rather than prescribed in advance.

Probing every candidate from the same nominal parameters makes the measured
responses directly comparable. The virtual parameters are discarded after
measurement, so only the recomposed direction is committed to the optimizer.
This finite-step construction evaluates the consequence of an update in the
units used by training, without requiring an analytic derivative through the
adversarial-example or weight-perturbation construction.

The measured dynamics reveal a persistent negative clean coefficient. A
majority of mini-batches already select $\beta_{\mathrm{clean}}<0$ at the
beginning of training, the proportion rapidly approaches one, and some early
mini-batches remain positive. This behavior is stronger than simply assigning
the clean direction a small weight: it crosses the zero-clean boundary into
explicit subtraction within the selected affine representation. A controlled
intervention confirms that setting $\beta_{\mathrm{clean}}=0$ does not
reproduce the same robustness profile as a mild negative value.

The zero point therefore marks the boundary between clean-gradient removal and
explicit affine subtraction. A positive coefficient retains clean descent, zero
removes it from the affine representation, and a negative coefficient
explicitly subtracts it. In the intervention, the adversarial and AWP
coefficients remain response-adaptive, isolating whether suppression is
sufficient or crossing into subtraction helps.

The affine coordinate must nevertheless be distinguished from the complete
update geometry. Because $\gadv$ and $\gawp$ can also project onto $\gclean$, a
negative $\beta_{\mathrm{clean}}$ does not imply that $\gbeta$ has a
negative net component along the clean axis. We therefore analyze
\begin{equation*}
p_{\mathrm{clean}}
=
\frac{\langle\gbeta,\gclean\rangle}{\|\gclean\|_2^2}.
\end{equation*}
Both $\beta_{\mathrm{clean}}$ and $p_{\mathrm{clean}}$ decrease as $R$
grows, indicating that stronger negative clean coefficients progressively
suppress the realized clean-axis component. Complete reversal is rarer and
appears mainly at the largest radius late in training. Norm-matched controls
further show that the directional effect survives after matching the
recomposed update to the AWP gradient norm, although update magnitude
contributes to part of the gain at larger $R$.

We evaluate ExGrad on CIFAR-10, CIFAR-100, and ImageNet-100 with two Wide
ResNet capacities. At suitable radii, ExGrad improves the principal robustness
metrics over AWP across all three datasets. The improvements include all four
robustness metrics on CIFAR-10 with WRN-34-10, all five reported clean and
robust metrics on CIFAR-100, and all four robustness metrics on ImageNet-100.
Our contributions are:
\begin{itemize}
    \item We separate source-loss descent sensitivity from finite-step
    adversarial-margin response and use the latter to select robust-training
    updates.
    \item We propose a lightweight, convex, batch-wise coefficient solver over
    the bounded affine span of clean, adversarial-input, and AWP gradients.
    \item We identify explicit clean-gradient subtraction, intervene across
    its zero boundary, and distinguish its affine coordinate from the net
    clean-axis projection.
    \item We use multi-seed cross-dataset evaluation and norm-matched controls
    to establish the performance effect and delimit scale-based explanations.
\end{itemize}

\ExGradOverviewFigure

\section{Related Work}

\paragraph{Robust objectives and instance regulation.}
PGD-AT approximates input-space robust optimization
\cite{madry2018towards}, while TRADES regulates natural and boundary errors
through an objective-level coefficient \cite{zhang2019theoretically}.
Robust generalization can require additional data
\cite{schmidt2018data,carmon2019unlabeled}, while standard learning may
exploit predictive but non-robust features \cite{ilyas2019features}.
MART, fast training, consistency regularization, and robust-overfitting studies
modify the objective, attack construction, regularization, or schedule
\cite{wang2020mart,wong2020fast,rice2020overfitting,tack2022consistency};
geometry-aware, margin-aware, and hard-class-aware methods reweight examples
\cite{zhang2021gairat,wang2021probabilistic,hou2023selfpaced}.
Together, these works expose the clean--robust trade-off
\cite{tsipras2019robustness}, but regulate how source losses generate an
update. ExGrad instead measures how already-constructed candidate updates
change the current adversarial margin.

\paragraph{Parameter-neighborhood robustness.}
AWP constructs a worst-case perturbation in a constrained weight neighborhood
and updates the nominal model using the perturbed-weight adversarial gradient
\cite{wu2020adversarial}. This complements the input neighborhood of PGD-AT
with parameter-neighborhood information. Sharpness-aware minimization provides
a broader parameter-neighborhood perspective \cite{foret2021sam}, and
robust-loss flatness has been linked to robust generalization
\cite{stutz2021flat}. Analyses of SAM further emphasize optimization-induced
implicit bias \cite{andriushchenko2022understanding}. ExGrad retains the
standard AWP state as a candidate rather than redefining its objective or
prescribing its weight.

\paragraph{Gradient recomposition and conflict.}
Multi-objective optimization and multi-task learning have developed
controls based on Pareto stationarity, rate balancing, conflict projection,
and worst-task improvement
\cite{sener2018multitask,chen2018gradnorm,yu2020pcgrad,liu2021cagrad}.
Conflict-aware adversarial training similarly studies interference between
clean and adversarial gradients \cite{xue2024conflictaware}. These methods
use source-gradient geometry or joint descent conditions. ExGrad instead
evaluates different input--parameter states against one margin target, permits
negative coefficients, and does not require every source loss to descend.

\paragraph{Virtual updates and margin signals.}
Virtual optimization steps have been used to learn example weights from a
validation objective
\cite{ren2018reweight,shu2019metaweight}. Task-affinity methods similarly
measure how one task's step changes another task's loss
\cite{fifty2021taskgrouping}. Margin-based attacks and training objectives
characterize robustness through shortest input-space distances to a decision
boundary \cite{moosavi2016deepfool,ding2020mma} or class-logit margins within
a perturbation set \cite{gao2022minimum}.
Without a validation set, ExGrad applies this response logic to restored probes
of the same mini-batch margin, then solves a low-dimensional affine problem.
Evaluation with PGD, APGD-CE, APGD-T, and AutoAttack guards against conclusions
tied to a weak or single attack \cite{athalye2018obfuscated,croce2020reliable}.

\section{Method}

\subsection{Candidate Gradients from Nested Training States}

Let $f_\theta$ be a classifier and $\ell$ cross-entropy. For each mini-batch,
we first generate detached adversarial examples
$x_j^{\mathrm{adv}}=x_j+\delta_j^\star$ using PGD at the nominal parameters
$\theta$. Holding these inputs fixed, AWP computes a detached perturbation
$\nu^\star$ with the standard layer-wise relative bound
$\|\nu_\ell^\star\|_2\le\gamma\|\theta_\ell\|_2$
\cite{wu2020adversarial}.
We define three source losses:
\begin{equation}
\left\{
\begin{aligned}
L_{\mathrm{clean}}
&=\frac{1}{B}\sum_{j=1}^{B}\ell(f_\theta(x_j),y_j),\\
L_{\mathrm{adv}}
&=\frac{1}{B}\sum_{j=1}^{B}\ell(f_\theta(x_j^{\mathrm{adv}}),y_j),\\
L_{\mathrm{awp}}
&=\frac{1}{B}\sum_{j=1}^{B}
\ell(f_{\theta+\nu^\star}(x_j^{\mathrm{adv}}),y_j).
\end{aligned}
\right.
\label{eq:candidates}
\end{equation}
Their candidate gradients are
$\gclean=\nabla_\theta L_{\mathrm{clean}}$,
$\gadv=\nabla_\theta L_{\mathrm{adv}}$, and
$\gawp=\nabla_\theta L_{\mathrm{awp}}$.
Both perturbations are treated as constants during these gradient
computations. After $\gawp$ is obtained, $\nu^\star$ is removed, so all probes
and the eventual optimizer update act on the same nominal parameters.

The candidates correspond to the nested state path
\begin{equation*}
(\theta,x)
\rightarrow
(\theta,x+\delta^\star)
\rightarrow
(\theta+\nu^\star,x+\delta^\star).
\end{equation*}

\subsection{Finite-Step Adversarial-Margin Response}

For the fixed adversarial input $x_j^{\mathrm{adv}}$, let $z_j(\theta)$ be its
logits. We use the signed top-1 margin and a detached sample-wise logit scale:
\begin{equation}
\left\{
\begin{aligned}
m_j(\theta)
&=
z_{j,y_j}(\theta)-\max_{k\ne y_j}z_{j,k}(\theta),\\
s_j
&=
\operatorname{Std}_k(z_{j,k}(\theta))+\varepsilon_s.
\end{aligned}
\right.
\label{eq:margin-scale}
\end{equation}
For each $i\in\{\mathrm{clean},\mathrm{adv},\mathrm{awp}\}$, ExGrad starts
from the same $\theta$, performs a virtual move
$\widetilde\theta_i=\theta-\eta_p\mathbf{g}_i$, measures the margin, and restores
$\theta$. The sample-wise response rate is
\begin{equation}
\left\{
\begin{aligned}
D_{j,i}
&=
\frac{m_j(\theta-\eta_p\mathbf{g}_i)-m_j(\theta)}{\eta_p},\\
\eta_p&=\kappa\eta.
\end{aligned}
\right.
\label{eq:probe-response}
\end{equation}
Here, $\eta$ is the optimizer learning rate and $\kappa$ is the probe factor.
Stacking the responses gives $\Dmat\in\mathbb R^{B\times3}$. For sufficiently
small $\eta_p$,
\begin{equation*}
D_{j,i}
\approx
-\langle\nabla_\theta m_j(\theta),\mathbf{g}_i\rangle.
\end{equation*}
The probe therefore estimates the response of the target margin to the
candidate parameter move rather than comparing source losses or candidate
angles. Parameter restoration also avoids differentiating through PGD, AWP,
or the candidate gradient fields.

The local affine prediction under $\betavec$ is
\begin{equation}
\widehat m_j(\betavec)
=
m_j(\theta)+\eta_p\Dmat_{j,:}\betavec.
\label{eq:predicted-margin}
\end{equation}
Let
\begin{equation*}
\begin{aligned}
b_j=\frac{m_j(\theta)}{s_j},
\qquad
\avec_j=\frac{\eta_p\Dmat_{j,:}^{\top}}{s_j}.
\end{aligned}
\end{equation*}
The normalized predicted margin is then
$r_j(\betavec)=b_j+\avec_j^\top\betavec$. The scale $s_j$ prevents examples with
large logit magnitudes from dominating the coefficient problem and remains
fixed throughout the local solve.

\subsection{Bounded Affine Coefficient Solver}

Let $\sigma(r)=(1+e^{-r})^{-1}$ denote the logistic sigmoid. We use the
log-sigmoid utility $u(r)=\log\sigma(r)$. Its derivative
$u'(r)=\sigma(-r)$ gives larger weight to adversarial examples
with small or negative predicted margins while saturating on already safe
examples. ExGrad solves
\begin{equation}
\left\{
\begin{aligned}
\betavec^\star
&=
\argmax_{\betavec\in\mathcal D} J(\betavec),\\
J(\betavec)
&=
\frac{1}{B}\sum_{j=1}^{B}
\log\sigma(b_j+\avec_j^\top\betavec),\\
\mathcal D
&=
\left\{\betavec\in\mathbb R^3\ \middle|\ 
\one^\top\betavec=1,\ \|\betavec\|_2\le R\right\}.
\end{aligned}
\right.
\label{eq:solver}
\end{equation}
The feasible set intersects the nonnegative simplex but does not impose
$\beta_i\ge0$. Writing $\bar\betavec=\one/3$, the unit-sum condition yields
\begin{equation*}
\|\betavec\|_2^2
=
\frac13+\|\betavec-\bar\betavec\|_2^2.
\end{equation*}
Hence $\mathcal D$ is a disk of radius
$\rho_R=\sqrt{R^2-1/3}$ within the unit-sum plane and is nonempty exactly when
$R\ge1/\sqrt3$. The radius controls how far the solver may move from the
uniform affine combination.

For fixed probe responses,
\begin{equation}
\nabla_{\betavec}^2J(\betavec)
=
-\frac{1}{B}\sum_{j=1}^{B}
\sigma(r_j)\sigma(-r_j)\avec_j\avec_j^\top
\preceq0.
\label{eq:hessian}
\end{equation}
The objective is therefore concave, while $\mathcal D$ is convex and compact.
The three-variable subproblem has a global optimum, and every KKT point is
globally optimal. This statement concerns only the local coefficient problem,
not neural-network training as a whole. In particular, the guarantee applies
to the finite-step affine response predictor in
Eq.~\eqref{eq:predicted-margin}, rather than the exact margin after the
recomposed update. We solve it by deterministic projected gradient ascent
with monotonic backtracking.

\subsection{Training Procedure and Geometric Diagnostics}

Each iteration performs five operations: generate $x^{\mathrm{adv}}$ and
$\nu^\star$; compute the three detached candidate gradients; remove
$\nu^\star$ and probe each direction from the same nominal parameters; solve
Eq.~\eqref{eq:solver}; and pass $\mathbf{g}_{\betavec^\star}$ to the optimizer for one
real update. Batch-normalization statistics and optimizer momentum are not
advanced during probes.

The coefficient $\beta_{\mathrm{clean}}$ is the explicit clean coordinate in
the selected affine frame. To account for correlations among candidates, we
also measure the clean-axis projection
\begin{equation}
p_{\mathrm{clean}}
=
\frac{\langle \gbeta,\gclean\rangle}{\|\gclean\|_2^2}.
\label{eq:projection}
\end{equation}
Expanding $\gbeta$ gives
\begin{equation*}
\begin{aligned}
p_{\mathrm{clean}}
&=
\beta_{\mathrm{clean}}
+\beta_{\mathrm{adv}}
\frac{\langle\gadv,\gclean\rangle}{\|\gclean\|_2^2}\\
&\quad+\beta_{\mathrm{awp}}
\frac{\langle\gawp,\gclean\rangle}{\|\gclean\|_2^2}.
\end{aligned}
\end{equation*}
A negative coefficient records explicit subtraction in the affine
representation, whereas $p_{\mathrm{clean}}<0$ records a negative net component
of the full update along the clean axis.

Relative to AWP, ExGrad reuses the PGD examples, weight-perturbation
construction, and the perturbed-weight adversarial backward pass. It
additionally computes clean and nominal-adversarial backward passes; the
remaining overhead consists of three margin-forward probes and a negligible
three-variable solve. ExGrad
does not change model size or inference cost.

\ExGradMainResultsTable

\section{Experiments}

\subsection{Experimental Setup}

\paragraph{Datasets and architectures.}
To test transfer across dataset scale and class diversity, we evaluate the
standard CIFAR-10 and CIFAR-100 adversarial-training benchmarks
\cite{krizhevsky2009learning} and ImageNet-100, a higher-resolution subset
constructed by randomly sampling 100 ImageNet classes
\cite{deng2009imagenet}. The sampled class list and further implementation
details are included in the supplementary code. WRN-28-4 is the principal
architecture, and CIFAR-10 additionally uses WRN-34-10
\cite{zagoruyko2016wide} to test transfer across capacity. All models are
trained for 200 epochs. CIFAR augmentation uses random cropping with
four-pixel padding and random horizontal flipping. ImageNet-100 uses random
resized crops and horizontal flips for training and center crops for
evaluation.

\paragraph{Optimization and attacks.}
All methods use SGD with learning rate $0.1$, momentum $0.9$, and weight decay
$5\times10^{-4}$, reducing the learning rate tenfold at epochs 100 and 150.
CIFAR uses PGD-10 training and PGD-20 evaluation with $\epsilon=8/255$ and step
size $2/255$; ImageNet-100 uses PGD-2 training and PGD-10 evaluation with
$\epsilon=4/255$. TRADES reports $\beta\in\{3,6,9\}$ under the same training
schedule and checkpoint protocol. AWP sets $\gamma=0.01$. ExGrad uses
$\kappa=0.1$; its projected solver is initialized at the uniform vector
$(1/3,1/3,1/3)$, runs for at most 50 iterations, and uses tolerance $10^{-6}$.

\ExGradTrainingCurvesFigure

\paragraph{Evaluation protocol.}
We report clean accuracy, PGD accuracy, APGD-CE, APGD-T, and standard
AutoAttack accuracy \cite{croce2020reliable}. ``PGD'' in
Table~\ref{tab:main-results} denotes PGD-20 on CIFAR and PGD-10 on
ImageNet-100. For each run, we select the checkpoint with the highest PGD
accuracy on the test set and evaluate every reported metric at that checkpoint.
This best-PGD rule is applied identically to all methods. Unless stated
otherwise, all tabulated accuracy results use matched seeds 0--2 and are
reported as mean $\pm$ sample standard deviation. Training dynamics and
projection diagnostics use seed 0.

\subsection{Main Robustness Results}

Table~\ref{tab:main-results} separates the role of the affine radius from
method comparisons. TRADES traces a clean--robust shift as $\beta$ increases
from $3$ to $9$: clean accuracy decreases from $84.94\%$ to $82.40\%$, while
PGD-20 increases from $50.23\%$ to $53.12\%$. On CIFAR-10 with WRN-28-4,
ExGrad at $R=0.8$ gives the most balanced point relative to AWP: it improves
clean, PGD-20, APGD-T, and AutoAttack accuracy, while APGD-CE is $0.23$ points
lower. Increasing $R$ to $1.0$ raises PGD-20 and APGD-CE by $2.47$ and $1.84$
points over AWP, respectively, but reduces clean and targeted robustness. At
$R=1.2$, PGD-20 reaches $59.23\%$, while clean and AutoAttack accuracy decline
further.

The larger WRN-34-10 model reproduces this radius-dependent pattern. At
$R=0.8$, ExGrad improves all four robustness metrics over AWP by $0.45$,
$0.59$, $0.48$, and $0.47$ points for PGD-20, APGD-CE, APGD-T, and
AutoAttack. At $R=1.0$, the PGD-20 and APGD-CE gains increase to $3.25$ and
$2.65$ points, whereas APGD-T and AutoAttack favor the smaller radius.

CIFAR-100 provides the most uniform improvement. At $R=0.8$, ExGrad exceeds
AWP on all five metrics, with gains of $3.84$, $1.43$, $1.12$, $1.72$, and
$1.75$ points in clean, PGD-20, APGD-CE, APGD-T, and AutoAttack accuracy.
At $R=1.0$, all four robustness metrics rise further and clean accuracy remains
$0.36$ points above AWP. Even $R=1.2$ retains gains on every robustness metric,
although clean accuracy falls below AWP.

On ImageNet-100, $R=0.8$ outperforms both PGD-AT and AWP on all four robust
metrics. Relative to AWP, the gains are $1.72$, $1.79$, $2.14$, and $2.15$
points for PGD-10, APGD-CE, APGD-T, and AutoAttack. Relative to PGD-AT, the
corresponding gains are $1.50$, $1.62$, $0.56$, and $0.55$ points. The
$R=1.0$ setting also exceeds both baselines on all four robust metrics, while
$R=1.2$ remains above AWP but not PGD-AT on APGD-T and AutoAttack.
The radius is therefore a bound on recomposition, not a universally monotonic
robustness control.

\subsection{Training Dynamics Against Baselines}

Figure~\ref{fig:training-curves} provides training-process context for the main
comparison. PGD-AT undergoes a pronounced PGD-20 decline after epoch 100,
whereas AWP substantially reduces this robust overfitting. ExGrad follows the
same stabilized regime but maintains higher PGD-20 accuracy through most of
the second half of training. Its clean trajectory remains below AWP,
consistent with the radius-dependent operating point in
Table~\ref{tab:main-results}. Overall, ExGrad provides sustained robustness
gains and stable training after the learning-rate decay.

\ExGradDynamicsFigure

\subsection{Explicit and Effective Clean-Gradient Subtraction}

Figure~\ref{fig:dynamics} contrasts the explicit clean coordinate with the
realized clean-axis projection at $R=0.8$, $1.0$, and $1.2$. At every radius,
the mean clean coefficient becomes negative early and its magnitude increases
with $R$. The coefficient dynamics therefore reveal increasingly strong
coordinate-level clean-gradient subtraction as the affine radius expands.

The sign event is also consistent across three seeds. Averaged over the three
seeds, $58.4\%$ of mini-batches already satisfy
$\beta_{\mathrm{clean}}<0$ at epoch 1 for $R=0.8$, and the fraction rapidly
approaches $100\%$. The same broad pattern holds for $R=1.0$ and $1.2$. Explicit
coordinate-level subtraction is therefore not restricted to a late or isolated
optimization phase.

The projection columns tell a different story. The realized
$p_{\mathrm{clean}}$ remains strongly positive at $R=0.8$, decreases
substantially at $R=1.0$, and approaches the zero boundary late in the
$R=1.2$ run.
Over epochs 151--200, the mean $p_{\mathrm{clean}}$ is $0.732$, $0.270$, and
$0.145$ for $R=0.8$, $1.0$, and $1.2$, while the proportions of negative
projections are $0.00\%$, $0.92\%$, and $22.28\%$. Thus a negative affine
coordinate can coexist with a positive net clean-axis component; effective
clean-axis reversal emerges mainly when the larger radius permits stronger
recomposition late in training.

\ExGradInterventionTable

\subsection{Candidate Complementarity and the Zero-Clean Boundary}

Observing a negative coefficient does not by itself show that its sign matters
for performance. Table~\ref{tab:fixed-clean} first removes each candidate in
turn while solving dynamically over the remaining pair. AT $+$ AWP and Clean
$+$ AWP reach similar mean PGD-20 accuracy, $56.44\%$ and $56.39\%$,
respectively. Removing AWP is substantially more damaging: Clean $+$ AT
reaches only $53.37\%$, roughly three points below the other pairs. None of
the three response-adaptive candidate-removal variants approaches full ExGrad
at $58.95\%$.

The AT $+$ AWP row also defines the zero-clean boundary. With the other two
coefficients response-adaptive, moving from $\beta_{\mathrm{clean}}=0.1$ to
$0$, $-0.1$, and $-0.2$ monotonically raises mean PGD-20 from $55.53\%$ to
$56.44\%$, $57.12\%$, and $57.40\%$, respectively, while clean accuracy
decreases. Removing the clean gradient is therefore not equivalent to
subtracting it, and no tested pair reproduces the full update.

Fixing the coefficients to their mean at the three seeds' best-PGD
checkpoints reaches $58.35\%$ PGD-20, $0.60$ points below
response-adaptive ExGrad. Thus the negative-clean affine region remains
useful, but one global composition does not fully replace batch-wise
response measurement.

\ExGradNormControlTable

\subsection{Direction Versus Update Magnitude}

Changing $R$ can alter both the direction and norm of $\gbeta$. We
construct a mini-batch-wise control
\[
\widetilde{\mathbf{g}}_{\betavec}
=
\frac{\|\gawp\|_2}{\|\gbeta\|_2+\varepsilon}
\gbeta,
\]
which preserves direction while matching the AWP gradient norm.
Matching changes mean PGD-20 by $+0.07$, $-0.30$, and $-0.37$ points at
$R=0.8$, $1.0$, and $1.2$. From $R=0.8$ to $1.2$, matched PGD-20 still rises
from $56.58\%$ to $58.86\%$, while clean accuracy falls from $83.81\%$ to
$79.92\%$. Update magnitude therefore does not explain the moderate-radius
gain or the clean--PGD operating-point shift.

\subsection{Training Cost}

In the single-H100 diagnostic-free benchmark of
Table~\ref{tab:training-cost}, timing includes data transfer, PGD-10, and all
required updates but excludes evaluation and I/O. ExGrad reuses AWP's
adversarial examples and weight-perturbation computation; it requires
$1.45\times$ AWP's epoch time and 24.24 MiB more peak allocated memory, with
no inference overhead. These measurements characterize controlled per-epoch
cost.

\ExGradTrainingCostTable

\section{Discussion}

Cross-zero and fixed-composition controls separate clean-gradient suppression,
explicit subtraction, and response adaptation. PGD-20 rises as
$\beta_{\mathrm{clean}}$ moves from $0.1$ through zero to $-0.2$, while the
fixed mean triplet recovers much, but not all, of the adaptive gain. This
reflects a response-preferred joint composition and a clean--robust
operating-point shift, not evidence that clean fitting is globally harmful;
nonnegative weighting cannot express this cross-zero regime.

Coefficient and projection signs describe different geometries: candidate
correlation means that $\beta_{\mathrm{clean}}<0$ need not imply
$p_{\mathrm{clean}}<0$. Updates at $R=0.8$ and $1.0$ usually retain a positive
clean-axis projection; only some late $R=1.2$ mini-batches reverse it. ExGrad
therefore establishes coordinate subtraction broadly and complete clean-axis
reversal more selectively.

The response objective is attack dependent: larger $R$ favors PGD and APGD-CE,
while moderate radii transfer better to APGD-T and AutoAttack. The local oracle
also omits momentum trajectories, targeted-attack candidates, and alternative
parameter-neighborhood directions. Multi-attack probes, cheaper response
estimators, and broader candidate states are natural extensions.

\section{Conclusion}

ExGrad selects a bounded affine gradient recomposition from finite-step
adversarial-margin responses. The solver repeatedly selects a negative clean
coefficient; cross-zero controls show that removal is insufficient and mild
subtraction helps when the adversarial-input and AWP directions remain.
Suitable radii improve robustness over AWP on CIFAR-10, CIFAR-100, and
ImageNet-100. Increasing $R$ lowers both $\beta_{\mathrm{clean}}$ and
$p_{\mathrm{clean}}$, suppressing the realized clean-axis component and
shifting the clean--PGD operating point.
Together, these results establish clean-gradient subtraction as a measured,
not prescribed, update preference. They expose an update-level geometry that
invites broader searches over robust training states.

\paragraph{Acknowledgments.}
This work was supported by the National Natural Science Foundation of China
(U22B2063) and the National Science and Technology Innovation 2030 Project of
China (2021ZD0202600). The model training was performed on the robotic
AI-Scientist platform of the Chinese Academy of Sciences.

\paragraph{Use of Generative AI.}
Generative AI tools were used for language editing and code assistance.
All technical content and experimental results were independently verified
by the authors.

\bibliography{exgrad_refs}

@inproceedings{szegedy2014intriguing,
  title = {Intriguing Properties of Neural Networks},
  author = {Szegedy, Christian and Zaremba, Wojciech and Sutskever, Ilya and Bruna, Joan and Erhan, Dumitru and Goodfellow, Ian and Fergus, Rob},
  booktitle = {International Conference on Learning Representations},
  year = {2014}
}

@inproceedings{goodfellow2015explaining,
  title = {Explaining and Harnessing Adversarial Examples},
  author = {Goodfellow, Ian J. and Shlens, Jonathon and Szegedy, Christian},
  booktitle = {International Conference on Learning Representations},
  year = {2015}
}

@inproceedings{madry2018towards,
  title = {Towards Deep Learning Models Resistant to Adversarial Attacks},
  author = {Madry, Aleksander and Makelov, Aleksandar and Schmidt, Ludwig and Tsipras, Dimitris and Vladu, Adrian},
  booktitle = {International Conference on Learning Representations},
  year = {2018}
}

@inproceedings{tsipras2019robustness,
  title = {Robustness May Be at Odds with Accuracy},
  author = {Tsipras, Dimitris and Santurkar, Shibani and Engstrom, Logan and Turner, Alexander and Madry, Aleksander},
  booktitle = {International Conference on Learning Representations},
  year = {2019}
}

@inproceedings{zhang2019theoretically,
  title = {Theoretically Principled Trade-off between Robustness and Accuracy},
  author = {Zhang, Hongyang and Yu, Yaodong and Jiao, Jiantao and Xing, Eric P. and El Ghaoui, Laurent and Jordan, Michael I.},
  booktitle = {International Conference on Machine Learning},
  pages = {7472--7482},
  year = {2019}
}

@inproceedings{rice2020overfitting,
  title = {Overfitting in Adversarially Robust Deep Learning},
  author = {Rice, Leslie and Wong, Eric and Kolter, J. Zico},
  booktitle = {International Conference on Machine Learning},
  pages = {8093--8104},
  year = {2020}
}

@inproceedings{wong2020fast,
  title = {Fast Is Better Than Free: Revisiting Adversarial Training},
  author = {Wong, Eric and Rice, Leslie and Kolter, J. Zico},
  booktitle = {International Conference on Learning Representations},
  year = {2020}
}

@inproceedings{wang2020mart,
  title = {Improving Adversarial Robustness Requires Revisiting Misclassified Examples},
  author = {Wang, Yisen and Zou, Difan and Yi, Jinfeng and Bailey, James and Ma, Xingjun and Gu, Quanquan},
  booktitle = {International Conference on Learning Representations},
  year = {2020}
}

@inproceedings{wu2020adversarial,
  title = {Adversarial Weight Perturbation Helps Robust Generalization},
  author = {Wu, Dongxian and Xia, Shu-Tao and Wang, Yisen},
  booktitle = {Advances in Neural Information Processing Systems},
  volume = {33},
  pages = {2958--2969},
  year = {2020}
}

@inproceedings{foret2021sam,
  title = {Sharpness-Aware Minimization for Efficiently Improving Generalization},
  author = {Foret, Pierre and Kleiner, Ariel and Mobahi, Hossein and Neyshabur, Behnam},
  booktitle = {International Conference on Learning Representations},
  year = {2021}
}

@inproceedings{andriushchenko2022understanding,
  title = {Towards Understanding Sharpness-Aware Minimization},
  author = {Andriushchenko, Maksym and Flammarion, Nicolas},
  booktitle = {International Conference on Machine Learning},
  volume = {162},
  pages = {639--668},
  year = {2022}
}

@inproceedings{sener2018multitask,
  title = {Multi-Task Learning as Multi-Objective Optimization},
  author = {Sener, Ozan and Koltun, Vladlen},
  booktitle = {Advances in Neural Information Processing Systems},
  volume = {31},
  pages = {525--536},
  year = {2018}
}

@inproceedings{chen2018gradnorm,
  title = {GradNorm: Gradient Normalization for Adaptive Loss Balancing in Deep Multitask Networks},
  author = {Chen, Zhao and Badrinarayanan, Vijay and Lee, Chen-Yu and Rabinovich, Andrew},
  booktitle = {International Conference on Machine Learning},
  pages = {794--803},
  year = {2018}
}

@inproceedings{yu2020pcgrad,
  title = {Gradient Surgery for Multi-Task Learning},
  author = {Yu, Tianhe and Kumar, Saurabh and Gupta, Abhishek and Levine, Sergey and Hausman, Karol and Finn, Chelsea},
  booktitle = {Advances in Neural Information Processing Systems},
  volume = {33},
  pages = {7164--7175},
  year = {2020}
}

@inproceedings{liu2021cagrad,
  title = {Conflict-Averse Gradient Descent for Multi-Task Learning},
  author = {Liu, Bo and Liu, Xingchao and Jin, Xiaojie and Stone, Peter and Liu, Qiang},
  booktitle = {Advances in Neural Information Processing Systems},
  volume = {34},
  pages = {18878--18890},
  year = {2021}
}

@inproceedings{ren2018reweight,
  title = {Learning to Reweight Examples for Robust Deep Learning},
  author = {Ren, Mengye and Zeng, Wenyuan and Yang, Bin and Urtasun, Raquel},
  booktitle = {International Conference on Machine Learning},
  pages = {4334--4343},
  year = {2018}
}

@article{xue2024conflictaware,
  title = {Conflict-Aware Adversarial Training},
  author = {Xue, Zhiyu and Wang, Haohan and Qin, Yao and Pedarsani, Ramtin},
  journal = {arXiv preprint arXiv:2410.16579},
  year = {2024}
}

@inproceedings{ding2020mma,
  title = {MMA Training: Direct Input Space Margin Maximization through Adversarial Training},
  author = {Ding, Gavin Weiguang and Sharma, Yash and Lui, Kry Yik Chau and Huang, Ruitong},
  booktitle = {International Conference on Learning Representations},
  year = {2020}
}

@inproceedings{athalye2018obfuscated,
  title = {Obfuscated Gradients Give a False Sense of Security: Circumventing Defenses to Adversarial Examples},
  author = {Athalye, Anish and Carlini, Nicholas and Wagner, David},
  booktitle = {International Conference on Machine Learning},
  pages = {274--283},
  year = {2018}
}

@inproceedings{croce2020reliable,
  title = {Reliable Evaluation of Adversarial Robustness with an Ensemble of Diverse Parameter-Free Attacks},
  author = {Croce, Francesco and Hein, Matthias},
  booktitle = {International Conference on Machine Learning},
  pages = {2206--2216},
  year = {2020}
}

@techreport{krizhevsky2009learning,
  title = {Learning Multiple Layers of Features from Tiny Images},
  author = {Krizhevsky, Alex and Hinton, Geoffrey},
  institution = {University of Toronto},
  year = {2009}
}

@inproceedings{deng2009imagenet,
  title = {{ImageNet}: A Large-Scale Hierarchical Image Database},
  author = {Deng, Jia and Dong, Wei and Socher, Richard and Li, Li-Jia and Li, Kai and Li, Fei-Fei},
  booktitle = {IEEE Conference on Computer Vision and Pattern Recognition},
  pages = {248--255},
  year = {2009}
}

@inproceedings{zagoruyko2016wide,
  title = {Wide Residual Networks},
  author = {Zagoruyko, Sergey and Komodakis, Nikos},
  booktitle = {British Machine Vision Conference},
  pages = {87.1--87.12},
  year = {2016}
}

@inproceedings{moosavi2016deepfool,
  title = {{DeepFool}: A Simple and Accurate Method to Fool Deep Neural Networks},
  author = {Moosavi-Dezfooli, Seyed-Mohsen and Fawzi, Alhussein and Frossard, Pascal},
  booktitle = {IEEE Conference on Computer Vision and Pattern Recognition},
  pages = {2574--2582},
  year = {2016}
}

@inproceedings{qin2019local,
  title = {Adversarial Robustness through Local Linearization},
  author = {Qin, Chongli and Martens, James and Gowal, Sven and Krishnan, Dilip and Dvijotham, Krishnamurthy and Fawzi, Alhussein and De, Soham and Stanforth, Robert and Kohli, Pushmeet},
  booktitle = {Advances in Neural Information Processing Systems},
  volume = {32},
  pages = {13824--13833},
  year = {2019}
}

@inproceedings{gao2022minimum,
  title = {Fast and Reliable Evaluation of Adversarial Robustness with Minimum-Margin Attack},
  author = {Gao, Ruize and Wang, Jiongxiao and Zhou, Kaiwen and Liu, Feng and Xie, Binghui and Niu, Gang and Han, Bo and Cheng, James},
  booktitle = {International Conference on Machine Learning},
  pages = {7144--7163},
  year = {2022}
}

@inproceedings{schmidt2018data,
  title = {Adversarially Robust Generalization Requires More Data},
  author = {Schmidt, Ludwig and Santurkar, Shibani and Tsipras, Dimitris and Talwar, Kunal and Madry, Aleksander},
  booktitle = {Advances in Neural Information Processing Systems},
  volume = {31},
  pages = {5019--5031},
  year = {2018}
}

@inproceedings{ilyas2019features,
  title = {Adversarial Examples Are Not Bugs, They Are Features},
  author = {Ilyas, Andrew and Santurkar, Shibani and Tsipras, Dimitris and Engstrom, Logan and Tran, Brandon and Madry, Aleksander},
  booktitle = {Advances in Neural Information Processing Systems},
  volume = {32},
  pages = {125--136},
  year = {2019}
}

@inproceedings{carmon2019unlabeled,
  title = {Unlabeled Data Improves Adversarial Robustness},
  author = {Carmon, Yair and Raghunathan, Aditi and Schmidt, Ludwig and Duchi, John C. and Liang, Percy},
  booktitle = {Advances in Neural Information Processing Systems},
  volume = {32},
  pages = {11190--11201},
  year = {2019}
}

@inproceedings{zhang2021gairat,
  title = {Geometry-Aware Instance-Reweighted Adversarial Training},
  author = {Zhang, Jingfeng and Zhu, Jianing and Niu, Gang and Han, Bo and Sugiyama, Masashi and Kankanhalli, Mohan},
  booktitle = {International Conference on Learning Representations},
  year = {2021}
}

@inproceedings{wang2021probabilistic,
  title = {Probabilistic Margins for Instance Reweighting in Adversarial Training},
  author = {Wang, Qizhou and Liu, Feng and Han, Bo and Liu, Tongliang and Gong, Chen and Niu, Gang and Zhou, Mingyuan and Sugiyama, Masashi},
  booktitle = {Advances in Neural Information Processing Systems},
  volume = {34},
  pages = {23258--23269},
  year = {2021}
}

@inproceedings{hou2023selfpaced,
  title = {Improving Adversarial Robustness with Self-Paced Hard-Class Pair Reweighting},
  author = {Hou, Pengyue and Han, Jie and Li, Xingyu},
  booktitle = {Proceedings of the AAAI Conference on Artificial Intelligence},
  volume = {37},
  pages = {14883--14891},
  year = {2023}
}

@inproceedings{tack2022consistency,
  title = {Consistency Regularization for Adversarial Robustness},
  author = {Tack, Jihoon and Yu, Sihyun and Jeong, Jongheon and Kim, Minseon and Hwang, Sung Ju and Shin, Jinwoo},
  booktitle = {Proceedings of the AAAI Conference on Artificial Intelligence},
  volume = {36},
  pages = {8414--8422},
  year = {2022}
}

@inproceedings{stutz2021flat,
  title = {Relating Adversarially Robust Generalization to Flat Minima},
  author = {Stutz, David and Hein, Matthias and Schiele, Bernt},
  booktitle = {IEEE/CVF International Conference on Computer Vision},
  pages = {7807--7817},
  year = {2021}
}

@inproceedings{shu2019metaweight,
  title = {{Meta-Weight-Net}: Learning an Explicit Mapping for Sample Weighting},
  author = {Shu, Jun and Xie, Qi and Yi, Lixuan and Zhao, Qian and Zhou, Sanping and Xu, Zongben and Meng, Deyu},
  booktitle = {Advances in Neural Information Processing Systems},
  volume = {32},
  pages = {1917--1928},
  year = {2019}
}

@inproceedings{fifty2021taskgrouping,
  title = {Efficiently Identifying Task Groupings for Multi-Task Learning},
  author = {Fifty, Christopher and Amid, Ehsan and Zhao, Zhe and Yu, Tianhe and Anil, Rohan and Finn, Chelsea},
  booktitle = {Advances in Neural Information Processing Systems},
  volume = {34},
  pages = {27503--27516},
  year = {2021}
}

\end{document}